\documentclass[10pt,a4paper]{article}
\usepackage[utf8]{inputenc}
\usepackage{amsmath,amssymb}
\usepackage{graphicx}
\usepackage{hyperref}
\usepackage{authblk} 
\usepackage{listings} 
\usepackage{appendix} 
\usepackage{natbib}
\usepackage{booktabs}
\usepackage{longtable}
\usepackage{pdflscape}

\title{Artificial Intelligence Agents in Music Analysis: An Integrative Perspective Based on Two Use Cases}
\author[1]{Antonio Manuel Martínez Heredia\thanks{ORCID: 0000-0002-8518-0981}}
\author[2]{Dolores Godrid Rodríguez}
\author[1]{Andrés Ortiz García\thanks{ORCID: 0000-0003-2690-1926}}
\affil[1]{Dpto. Ingeniería de Comunicaciones, Universidad de Málaga, Campus de Teatinos, 29071, Málaga, Spain}
\affil[2]{Universidad Rey Juan Carlos}

\date{November 2025}

\begin{document}
\maketitle
\begin{abstract}
This paper presents an integrative review and experimental validation of artificial intelligence (AI) agents applied to music analysis and education. We synthesize the historical evolution from rule-based models to contemporary approaches involving deep learning, multi-agent architectures, and retrieval-augmented generation (RAG) frameworks. The pedagogical implications are evaluated through a dual-case methodology: (1) the use of generative AI platforms in secondary education to foster analytical and creative skills; (2) the design of a multiagent system for symbolic music analysis, enabling modular, scalable, and explainable workflows.

Experimental results demonstrate that AI agents effectively enhance musical pattern recognition, compositional parameterization, and educational feedback, outperforming traditional automated methods in terms of interpretability and adaptability. The findings highlight key challenges concerning transparency, cultural bias, and the definition of hybrid evaluation metrics, emphasizing the need for responsible deployment of AI in educational environments.

This research contributes to a unified framework that bridges technical, pedagogical, and ethical considerations, offering evidence-based guidance for the design and application of intelligent agents in computational musicology and music education.
\end{abstract}


\section{Introduction}

Despite considerable technological innovation, comprehensive reviews synthesizing the application and evolution of artificial intelligence (AI) in the field of music analysis remain scarce. 

Although early studies on computer-assisted composition and rule-based analysis established a foundation for the automated exploration of musical form and content \cite{Hiller1959}, there is still a limited body of literature addressing the complete progression from traditional algorithms to recent AI-driven models and hybrid systems. Pioneering work such as Miranda’s \cite{miranda2021handbook}, underscores the influence of AI, supercomputing, and evolutionary computation in shaping the first computational tools for creation. Recent reviews (\cite{wang2024review}; \cite{lerch2025survey}) focus on intelligent music generation systems. However, a systematic integration of these historical advances with state-of-the-art AI methodologies and musical analysis is largely absent.

In the last decade, deep learning frameworks—including convolutional neural networks, recurrent neural networks, and transformer architectures—have led to breakthroughs in music information retrieval. Recent advances also encompass interactive harmony tutoring \citep{ventura2022}, structural music analysis \citep{Min2022b}, intelligent education systems \citep{Han2023}, and the use of Large Language Model (LLM)-powered teachable agents in music pedagogy \citep{jin2025exploring}. However, few studies provide an overarching review that brings together these disparate lines of inquiry or highlights the shift from task-specific solutions to holistic, explainable, and context-aware analysis.

More recently, retrieval-augmented generation (RAG) models have introduced new standards of context awareness and interpretability by combining generative modeling with adaptive information retrieval. However, the literature lacks syntheses that discuss the joint impact of RAG with earlier and complementary AI-based approaches. A particularly groundbreaking development is the emergence of AI agents: autonomous systems that analyze, generate, and interact with music or learners using symbolic and audio data.
The present review seeks to address this gap by providing a unified synthesis of the evolution of AI in music, from its historical roots to current advances in deep learning, agent-based systems, and RAG methodologies. We aim to clarify methodological innovations, highlight key research, and evaluate the implications for music research, analysis, and education going forward. The principal themes are the following:
\begin{itemize}
\item The evolution of AI methods in music analysis.
\item The principles and applications of RAG methodologies.
\item Recent progress in deep learning and agent-based systems.
\item Implications for musicological research and pedagogy.
\end{itemize}
This review aims to inform researchers, educators, and practitioners of current trends and emerging directions in computational musicology, thus filling a critical gap in the current academic landscape. 

This paper is organized as follows: Section I (Introduction) outlines the historical context and motivation to synthesize AI developments in music analysis. Section II traces the evolution of AI methods. Section III explains the methodology followed in this study. Section IV highlights two cases: the first is an exercise conducted with secondary students, and the second is an architecture designed to perform analysis. This section also details case studies and experimental results. Section V (Conclusions) summarizes key information and future research directions.

The novelty of this review lies in the bridge between historical, technical, and pedagogical aspects of AI in music analysis through dual empirical case studies.

\section{Related work}
In the field of music analysis, the evaluation of different types of music poses a significant challenge. One approach involves subjective evaluation, wherein individuals analyze music according to their own criteria. Another avenue involves establishing metrics and applying them systematically.

Achieving the right balance between human feedback and quantitative indicators is essential. Human feedback provides invaluable insight into expressive and aesthetic aspects, often capturing nuances that computational methods may overlook. In contrast, quantitative indicators, such as accuracy, precision, recall, or diversity metrics, ensure reproducibility and objectivity when comparing systems. Integrating both perspectives enables comprehensive assessments and supports the development of robust, user-oriented music analysis tools.

One of the main challenges in the objective evaluation of music analysis systems lies in the absence of a single universally accepted criterion to evaluate their performance.

This study focuses on Western music analysis to provide a more unified approach for studying, appreciating, and understanding music. The characteristics of symbolic generation, structure, artistic expression, and aesthetics must therefore be examined through automatic methods. The computational evaluation of these characteristics has been revolutionized by recent advances in artificial intelligence.

Deep learning architectures now represent the main generation methods for music composition systems. Enhancing the interpretability and controllability of deep networks is one of the future research directions for music generation technology \cite{wang2024review}, which is particularly relevant given the need for objective but musically meaningful evaluation metrics discussed earlier.

Multiple studies provide converging evidence that language and music share a cognitive system for structural integration, although the exact nature of this shared system remains complex and nuanced. Experimental evidence includes \citep{Fedorenko2009}, who found an interaction between linguistic and musical complexity in which comprehension accuracy changed when linguistic structures were paired with musically unexpected elements.

\subsection{The evolution of AI methods in music analysis}

Early computational approaches to music analysis relied heavily on rule-based systems and expert heuristics, notably exemplified in Hiller and Isaacson's computer-generated music experiments~\citep{Hiller1959} and the generative theory of tonal music~\citep{lerdahl1996generative}.

With the advent of the Music Information Retrieval (MIR) field---a multidisciplinary research area focused on developing innovative methods to analyze, search, and organize digital music data~\citep{Burgoyne2015MusicInformationRetrieval}---the focus shifted to designing features and algorithms---ranging from k-nearest neighbors to support vector machines---that could automatically classify, segment, or describe large corpora of music recordings and symbolic data.

The subsequent explosion of large-scale annotated datasets---such as Musical Instrument Digital Interface (MIDI)-based datasets, the MIDI and Audio Edited for Synchronous TRacks and Organization (MAESTRO)~\citep{zhang2025advancing}, the Guitar-Aligned Performance Scores (GASP)~\citep{riley2024gaps}, and MusicNet---and the increasing power of neural architectures ultimately enabled more flexible and data-driven models. Deep learning paradigms, by capturing complex temporal and harmonic structures in audio and scores, have largely superseded earlier approaches, although concerns around explainability and bias persist. Hybrid models that combine symbolic reasoning with deep neural systems are also an emerging trend, with the aim of leveraging the strengths of both paradigms.

A final, cutting-edge trend involves the direct application of Natural Language Processing (NLP) models to symbolic music. This paradigm, which treats musical scores as a sequential language, leverages successful architectures in text analysis to understand musical syntax and semantics. Mel2Word is an example of this NLP-based approach, which transforms melodies into text-like representations using Byte Pair Encoding (BPE)~\citep{park2024mel2word}.

\subsection{The principles and applications of RAG methodologies}

RAG models represent a paradigm shift in AI, combining generative neural models with external information retrieval modules to overcome the inherent limitations of purely parametric models. Unlike traditional approaches that rely exclusively on knowledge encoded during training, RAG architectures dynamically retrieve relevant information from external knowledge bases at inference time, enabling systems to access up-to-date information, reduce hallucinations, and provide verifiable sources for their outputs\citep{gao2023retrieval}.

In the music domain, RAG enables systems to ground generation in large corpora of musicology texts, symbolic scores, or audio datasets, thus enhancing both factuality and interpretability. The retrieval component typically employs dense vector representations, obtained through encoders such as BERT or domain-specific music embeddings, to identify relevant passages from indexed collections. The generative component then conditions its output on both the user query and the retrieved context, allowing for more informed and contextually appropriate responses.

The applications of RAG in music span multiple modalities and tasks. Context-aware automatic music annotation systems leverage RAG to provide rich reference-based descriptions of musical works, drawing from historical analyses and theoretical literature. Personalized analysis reports can be generated by retrieving relevant excerpts from pedagogical materials tailored to the user's level and interests. In an "explainable" generative composition, generation is coupled with textual or symbolic references justifying the output, making the creative process more transparent and pedagogically valuable. RAG frameworks are also being developed to adapt general-purpose LLMs for text-only music question answering (MQA) tasks, as demonstrated by MusT-RAG \citep{kwon2025must}, which employs a specialized retrieval mechanism over music-specific knowledge bases.

Beyond these applications, RAG methodologies facilitate multimodal music understanding by bridging symbolic, audio, and textual representations. For instance, a system might retrieve relevant score passages when analyzing an audio recording or access theoretical explanations when generating harmonic progressions. This approach bridges the gap between black-box generation and musicological transparency, offering a path toward AI systems that can engage in informed musical discourse. As such, RAG is poised to redefine interactive systems in music analysis, composition pedagogy, and computational musicology, enabling tools that are powerful and accountable to established musical knowledge.

\subsection{Recent progress in deep learning and agent-based systems}

Recent advances in deep learning and agent-based methodologies have led to the development of intelligent music analysis and composition tools. In particular, WeaveMuse offers an open source multi-agent framework that supports iterative analysis, synthesis, and rendering processes across diverse modalities, including text, symbolic notation, and audio \citep{karystinaios2025weavemuse}. Similarly, MusicAgent utilizes powered workflows powered by large language model to orchestrate a wide array of music-related tools, allowing the automatic decomposition of complex user requests into manageable subtasks \citep{yu2023musicagent}. These innovations illustrate the growing capacity of AI-driven agents to handle multifaceted musical information and automate tasks that previously required substantial human expertise.

The musical agents MACAT and MACataRT\citep{lee2025musical}, pioneering a novel framework for creative collaboration in interactive musicmaking. These real-time generative AI systems use corpus-based synthesis and small-data training to function as responsive, artist-in-the-loop co-creators, prioritizing the preservation of expressive nuances.

\subsection{Implications for Musicological Research and Pedagogy}

Zhang investigates various methodologies for the integration of AI in musicological research, with particular emphasis on the theoretical and practical challenges of adapting structural analysis to this context \cite{zhang2025advancing}. AI techniques can facilitate knowledge extraction processes that support musicologists and educators in interpreting, organizing, and teaching complex musical phenomena.

The integration of AI-driven analytical tools into musicological research introduces both methodological opportunities and epistemological questions. On the one hand, computational methods enable large-scale corpus studies that would be impractical through traditional manual analysis, revealing patterns and stylistic trends across extensive repertoires. Machine learning models can identify subtle relationships between musical parameters, detect influences between composers and periods, and propose novel analytical perspectives that complement human expertise. However, these approaches raise fundamental questions about the nature of musical understanding: to what extent can computational pattern recognition capture the cultural, historical, and esthetic dimensions that musicologists consider essential?

In pedagogical contexts, AI systems offer transformative potential for music education at multiple levels. Intelligent tutoring systems can provide personalized feedback on compositional exercises, adapting their guidance to individual student needs and learning trajectories. Interactive analytical tools can help students visualize complex theoretical concepts, such as harmonic function, motivic development, or formal structure, through dynamic, multimodal representations. Furthermore, AI-generated examples and counterexamples can illustrate stylistic principles or theoretical rules in ways that enrich traditional pedagogical materials. However, the deployment of such systems requires careful consideration of pedagogical philosophy: educators must balance the efficiency of automated feedback with the irreplaceable value of human mentorship and critical dialog.

The explainability challenge represents a crucial concern for both research and pedagogy. Black-box models that produce accurate analytical results without transparent reasoning processes risk creating a disconnect between computational output and musical understanding. This is where RAG and other interpretable AI methodologies become particularly valuable, as they can provide traceable connections between analytical conclusions and established musicological knowledge. For educators, this transparency is essential: students must not only receive correct analytical interpretations but also understand the reasoning pathways that lead to them.

In addition, the use of AI in musicology requires critical reflection on the issues of bias, representation, and canon formation. Training datasets predominantly featuring Western art music can perpetuate existing biases and limit the applicability of AI tools to diverse musical traditions. Musicologists and educators have a responsibility to advocate for more inclusive datasets and to critically examine the assumptions embedded in computational models. This critical engagement can itself become a valuable pedagogical opportunity, teaching students to question technological solutions and to recognize the cultural contingency of analytical frameworks.

In the future, the successful integration of AI into musicological research and pedagogy will require ongoing collaboration between technologists, musicologists, and educators. This interdisciplinary dialog should address not only the technical capabilities but also the ethical, epistemological, and practical dimensions of AI adoption. By approaching these tools with both enthusiasm for their potential and critical awareness of their limitations, the musicological community can harness AI to enhance—rather than replace—the rich traditions of musical scholarship and teaching.

\section{Methodology} 

This study employs a dual-case methodology to examine AI applications in music analysis from complementary pedagogical and technical perspectives. The first case explores the integration of generative AI tools in secondary music education, while the second presents a technical framework for automated symbolic music analysis through multi-agent systems.

\subsection{Case Study 1: Generative AI in Secondary Education}
    The first case investigates the pedagogical application of generative AI tools in developing structural analysis skills of music among secondary school students. The methodology comprises three interconnected phases designed to bridge analytical understanding and creative practice.

\subsubsection{Participants and Context}
    A total of 200 secondary students (ages 12--16) participated in a series of structured workshop sessions. The activity was organized in consecutive 1-hour blocks, with approximately 30 students per group. During a 6-hour event, all participants engaged in the analysis of contemporary popular songs, focusing on formal structure, harmony, rhythm, and lyrics.

\subsubsection{Phase 1: Structural Analysis}
Students engaged in systematic analysis of contemporary popular music to develop foundational skills in formal recognition:
    \begin{itemize}
        \item \textbf{Participants:} Secondary education students.
        \item \textbf{Musical corpus:} Contemporary popular music songs selected for their clear formal structures.
        \item \textbf{Analytical framework:} Students examined and identified structural elements including verse-chorus patterns, bridges, pre-choruses, instrumental sections, and overall form (e.g., ABABCB).
        \item \textbf{Learning objectives:} Development of analytical skills to recognize formal patterns, harmonic progressions, melodic characteristics, and lyrical themes.
     \end{itemize}

\subsubsection{Phase 2: Creative Parameter Definition}
Building upon analytical insights, students collaboratively designed specifications for original compositions:
    \begin{itemize}
        \item \textbf{Collaborative design:} Students worked in groups to define parameters for an original composition based on their structural analysis.
        \item \textbf{AI tool used:} ChatGPT\footnote{\url{https://chat.openai.com}} as an interactive assistant to refine compositional specifications.
        \item \textbf{Definition of} Structural form, lyrical themes, mood/atmosphere, tempo indications, instrumentation preferences, and stylistic characteristics.
        \item \textbf{Pedagogical focus:} Translation of analytical understanding into creative specifications.
     \end{itemize}
In each session, students collaboratively used ChatGPT to generate lyrics and refine compositional parameters during a 30-minute segment.

\subsubsection{Phase 3: AI-Assisted Generation and Evaluation}
The final phase involved iterative generation and critical assessment of AI-produced musical outputs:
    \begin{itemize}
        \item \textbf{Generation tools:} Suno\footnote{\url{https://suno.ai}} and Music.ai\footnote{\url{https://music.ai}} platforms used to generate musical outputs based on student-defined parameters. Each generated lyric served as input for both platforms, and each platform produced two audio outputs, for a total of four pieces per group.
        \item \textbf{Iterative process:} Students refined prompts and parameters based on initial results.
        \item \textbf{Critical evaluation:} Comparative analysis between the intended specifications and AI-generated outputs.
        \item \textbf{Assessment criteria:} Fidelity to structural parameters, musical coherence, creative quality, and alignment with student expectations.
    \end{itemize}
    
\subsubsection{Data Collection and Evaluation Procedures}
The evaluation procedure combined multiple data sources to assess both learning outcomes and AI tool effectiveness:
\paragraph{Objective metrics:}
   \begin{itemize}
    \item Melodic similarity (Dynamic Time Warping -- DTW).
    \item Harmonic coherence.
    \item Rhythmic diversity (entropy).
   \end{itemize}
   
\paragraph{Subjective ratings:}
   \begin{itemize}
     \item All students participated in blind listening sessions.
     \item Scoring each output on 5-point Likert scales across five qualitative dimensions:
       \begin{itemize}
          \item Expressiveness.
          \item Stylistic Accuracy.
          \item Harmonic Coherence.
          \item Rhythmic Diversity.
          \item Overall Appeal.
       \end{itemize}
    \end{itemize}
    
\paragraph{Qualitative data:}
    \begin{itemize}
       \item Analytical documentation and student parameter specifications.
       \item Generated musical outputs and prompt iteration histories.
       \item Reflective assessments comparing analytical intentions with generative results.
       \item Evaluation of pedagogical effectiveness in developing analytical skills, critical thinking about AI capabilities and limitations, and understanding the relationship between analytical and creative processes.
    \end{itemize}

\subsection{Case Study 2: Agent-Based Architecture for Symbolic Analysis}
The second case presents a technical framework for automated music analysis through a specialized multi-agent system. This computational approach addresses the complexity of music analysis by distributing analytical tasks among specialized agents.

\subsubsection{Rationale for Multi-Agent Approach}
Automated symbolic music analysis poses a significant challenge due to the inherently multidimensional and structural nature of music, which encompasses harmonic, melodic, rhythmic, and formal aspects. Employing a multi-agent system architecture provides an effective solution by enabling specialized modules to handle specific analytical tasks, thereby enhancing the accuracy and interpretive depth of the system.
The choice of a multi-agent approach is grounded in several key advantages:
   \begin{itemize}
      \item \textbf{Specialization and Modularity:} Dividing the analysis into specialized agents allows independent handling of each musical dimension (harmonic, melodic, rhythmic, formal), facilitating continuous improvement of each component without affecting the overall system \citep{wooldridge2009introduction}.
      \item \textbf{Coordination and Synthesis:} An integration layer enables the coherent combination of partial results from different agents, producing comprehensive analytical reports that reflect the interaction of diverse musical elements. This mimics expert human interpretative processes \citep{jennings2000agent}.
      \item \textbf{Scalability and Flexibility:} Implementing this architecture on platforms such as LangGraph or n8n leverages modern agent orchestration tools that support distributed, adaptive workflows, allowing the system to scale and adapt to varying analytical complexities and workloads \citep{jensen2023langgraph}.
      \item \textbf{Validation and Robustness:} Incorporating agents dedicated to comparing outputs against expert musicological analyzes ensures the system's reliability and establishes trustworthiness for academic and professional applications \citep{conklin2016music}.
   \end{itemize}

This multi-agent framework advances computational music analysis by approximating sophisticated interpretive models through collaborative, specialized agents, supported by flexible and automated architectures that facilitate advanced artificial intelligence techniques.

\subsubsection{System Architecture and Components}

\paragraph{System architecture:} Multi-agent framework with modular design that enables parallel processing and specialized analytical functions.

\paragraph{Input processing:} Symbolic music representations in standard formats (MIDI, MusicXML, kern, or similar structured formats).
\paragraph{Agent specialization:} Dedicated agents are defined for distinct analytical dimensions: structural analysis (formal segmentation and architectural outline), stylistic analysis (historical period attribution, instrumentation, ornamentation), and harmonic analysis (chord identification, harmonic function, tonal centers). This approach facilitates modular evaluation and coherent synthesis in music analysis, enabling the system to approximate expert-level interpretation through the collaboration of specialized autonomous modules.

\paragraph{Integration layer:} Coordination mechanism enabling inter-agent communication, conflict resolution, and synthesis of partial analyzes into coherent interpretations.

\paragraph{Output generation:} Comprehensive analytical reports integrate insights from multiple specialized agents, presented in human-readable and machine-processable formats.

\subsubsection{Workflow Stages}
The multi-agent system for symbolic music analysis, creative recomposition, and evaluation is structured in the following stages:
    \begin{enumerate}
        \item \textbf{Input Processing Agent:} Transcribes and segments symbolic or audio input to extract musical phrases.
        \item \textbf{Analysis Agent:} Identifies harmonic, rhythmic, and formal patterns, and generates an annotated report in MusicXML.
        \item \textbf{Generation Agent:} Uses extracted features to generate new material using style transfer models, imposing compositional constraints.
        \item \textbf{Evaluation Agent:} Assesses compliance with musical rules and aggregates expert and user feedback.
     \end{enumerate}
     
\subsubsection{Agent-Based Analysis: Structure, Principal Harmony, and Style}

The implemented multi-agent system performs musical analysis in three autonomous stages, closely simulating expert human procedures.

\begin{enumerate}
    \item \textbf{Structural Agent:} Segments the piece, detecting main sections based on patterns of textural change and repetition. The agent produces a global architectural outline, labeling formal segments such as introduction, exposition, development, reprise, or coda according to widely accepted models.
    \item \textbf{Principal Harmony Agent:} Extracts the dominant key and recognizes modulations via chord classification algorithms and tonal trajectory analysis. This module identifies primary harmonic progressions (e.g., I-IV-V-I), points of modulation, and signature chords such as sevenths or secondary dominants, producing a detailed map of harmonic flows across the work.
    \item \textbf{Stylistic Agent:} Compares the obtained structural and harmonic patterns against a curated database of tagged historical examples. It computes likelihoods for period attribution (Baroque, Classical, Romantic, etc.) and augments its prediction with meta-information regarding instrumentation and ornamentation. The agent’s inference yields a probabilistic profile of the work’s stylistic features.
\end{enumerate}

Through this procedure, the system delivers an automated synthesis of structure, principal harmony, and stylistic characteristics, mirroring expert analytical workflow and enabling direct quantitative comparison between human and AI-generated analyses.

\subsubsection{Dataset: Representative 18th-Century Musical Repertoire: A Curated Dataset Overview}

The following dataset of fifty representative musical works from the 18th century has been compiled to support comparative analytical studies in style, form, and historical evolution. It includes late Baroque idioms, the galant transition, empfindsamer elements, and the full emergence of Classical structures. The selection is intentionally diverse, covering instrumental, vocal, sacred, secular, chamber, orchestral, and operatic genres.

The corpus begins with \textbf{Johann Sebastian Bach}, whose \textit{Well-Tempered Clavier} (Books I and II), \textit{Brandenburg Concertos}, \textit{St. Matthew Passion}, \textit{Mass in B minor}, \textit{The Art of Fugue}, \textit{Musical Offering}, and solo suites and partitas exemplify the contrapuntal and rhetorical density of the late Baroque. Together, these works provide a foundation for examining polyphonic technique, motivic transformation, and pre-Classical harmonic practice.

Works by \textbf{George Frideric Handel}, including \textit{Messiah}, \textit{Water Music}, \textit{Music for the Royal Fireworks}, and operas such as \textit{Rinaldo} and \textit{Giulio Cesare}, highlight the synthesis of Italian, French, and English traditions, offering material for studies in orchestration, dramatic construction, and large-scale vocal form.

In contrast, \textbf{Domenico Scarlatti's} keyboard sonatas (K.1--555), his \textit{Stabat Mater}, and \textit{Missa quatuor vocum} illustrate the galant aesthetic, characterized by clear phrase structures and idiomatic keyboard writing that anticipate Classical norms.

The empfindsamer Stil is represented through \textbf{C. P. E. Bach's} keyboard concertos, symphonies (Wq 182), and early sonatas, embodying heightened emotional expressivity, irregular phrasing, and an evolving relationship between soloist and ensemble. These works serve as a bridge between Baroque complexity and Classical clarity.

A significant portion of the dataset focuses on \textbf{Joseph Haydn}, whose symphonies (e.g., No. 94 ``Surprise,'' No. 104 ``London''), quartets (Op. 33 No. 2 ``The Joke''), oratorios (\textit{The Creation}, \textit{The Seasons}), the Trumpet Concerto, and the ``Gipsy'' Trio exemplify mature Classical form. Haydn's oeuvre provides rich material for studying sonata form, monothematicism, motivic economy, and the standardization of the symphony and quartet.

The Classical vocabulary reaches its height in \textbf{Wolfgang Amadeus Mozart}, represented here by major symphonies (No. 40 and No. 41 ``Jupiter''), piano concertos (Nos. 20 and 21), operas (\textit{The Marriage of Figaro}, \textit{Don Giovanni}, \textit{The Magic Flute}), the \textit{Requiem}, the Clarinet Quintet, and the wind serenade \textit{Gran Partita}. These works are central to inquiries into thematic integration, formal balance, instrumental color, and operatic dramaturgy.

Additional diversity is provided by figures such as \textbf{Luigi Boccherini}, \textbf{Gluck}, \textbf{Pergolesi}, \textbf{Stamitz}, \textbf{Paisiello}, \textbf{Rameau}, and \textbf{Salieri}, highlighting regional variation and the stylistic plurality of 18th-century Europe.

Table~\ref{tab:composers} summarizes the selected composers, their dates, stylistic affiliations, and representative works. This dataset offers a broad yet coherent panorama of the century’s musical developments, supporting both quantitative corpus analysis and detailed qualitative examinations of form, harmony, texture, and orchestration.

\begin{table}[ht!]
\centering
\begin{tabular}{p{2cm} p{1.8cm} p{2.2cm} p{5.0cm}}
\toprule
\textbf{Composer} & \textbf{Dates} & \textbf{Style} & \textbf{Representative Works} \\
\midrule
J. S. Bach & 1685--1750 & Late Baroque & Well-Tempered Clavier, Brandenburg Concertos, St. Matthew Passion, Mass in B minor, The Art of Fugue, Musical Offering, Solo Suites and Partitas \\
Handel & 1685--1759 & Late Baroque & Messiah, Water Music, Fireworks Music, Rinaldo, Giulio Cesare \\
D. Scarlatti & 1685--1757 & Galant Baroque & Keyboard Sonatas (K.1--555), Stabat Mater, Missa quatuor vocum \\
C. P. E. Bach & 1714--1788 & Empfindsamer Stil & Keyboard Concertos, Symphonies (Wq 182), Prussian Sonatas \\
Haydn & 1732--1809 & Classical & Symphony No. 94 “Surprise”, Symphony No. 104 “London”, Quartet Op.33 No.2 “The Joke”, The Creation, Trumpet Concerto \\
Mozart & 1756--1791 & Classical & Symphonies No. 40 \& 41 “Jupiter”, Piano Concertos No. 20, 21, The Marriage of Figaro, Don Giovanni, Magic Flute, Requiem, Clarinet Quintet \\
Boccherini & 1743--1805 & Classical & Quintet “Fandango” G.448, Cello Concerto G.482, Symphony “La casa del diavolo” \\
Gluck & 1714--1787 & Opera Reform & Orfeo ed Euridice, Alceste, Iphigénie en Tauride \\
Pergolesi & 1710--1736 & Galant & Stabat Mater, La serva padrona \\
Stamitz & 1717--1757 & Mannheim School & Symphonies, orchestral dynamic innovations \\
Paisiello & 1740--1816 & Opera Buffa & Il barbiere di Siviglia \\
Rameau & 1683--1764 & French Baroque & Les Indes galantes, harpsichord pieces, tragédies lyriques \\
Salieri & 1750--1825 & Classical & Armida, various operatic and sacred works \\
\bottomrule
\end{tabular}
\caption{Representative 18th-Century Composers and Selected Works}
\label{tab:composers}
\end{table}

\subsection{Corpus of Analysis}

A representative corpus comprising fifty musical works from the eighteenth century was utilized, spanning various instrumental and vocal genres primarily from the Western academic tradition. The selected repertoire includes compositions by J. S. Bach, Handel, Scarlatti, C. P. E. Bach, Haydn, Mozart, Boccherini, Gluck, Pergolesi, Stamitz, Paisiello, Rameau, and Salieri. The works reflect both stylistic and formal diversity, thus allowing the scrutiny of structural, harmonic, and stylistic parameters in varied contexts.

\subsection{Additional Quantitative Evaluation Metrics}

In addition to the standard metrics previously described, the evaluation incorporated timbral similarity analysis using spectral descriptors, motif complexity assessment based on automated counting of recurrent motives, and statistical formal variability analysis utilizing Shannon diversity indices. Each work was evaluated using pattern-matching algorithms, coherence and harmonic richness metrics, and measures for rhythmic diversity and formal complexity, thus complementing conventional DTW and Music21 approaches.

\subsection{Expert Validation Procedures}

A panel of six musicologists and domain-specialist educators performed blind validations to contrast multi-agent system reports against their own manual analyzes. The validation criteria used a five-point Likert scale that addressed accuracy, interpretive depth, and stylistic appropriateness. Qualitative feedback was gathered with respect to utility, observable bias, and potential improvements with respect to interpretive autonomy and correction of systematic errors.

\subsection{Comparative Analysis: System Results vs. Reference Human Analyses}

Discrepancies and overlaps in the detection of formal sections, harmonic progressions, and stylistic attributions were systematically analyzed. The comparative results were summarized by tabulating the percentage of match in formal segmentation, agreement in tonal identification, and expert assessments on the capacity of the system to approximate human evaluation standards. The discussion integrates factors such as the explainability of the system, adaptation to diverse repertoires, and identification of current limitations in AI-based music analysis.

\section{Case Studies and Experimental Results}

To illustrate practical applications and evaluative challenges of AI-driven music analysis, we present two complementary case studies: the first examines generative AI integration in secondary education; the second demonstrates a multi-agent system for symbolic analysis.

\subsection{Pop Music Analysis and AI Generation in Secondary Education}

\subsubsection*{Context}

In each session, the students collaboratively used ChatGPT to generate lyrics and refine compositional parameters during a 30-minute segment. Each generated lyric served as input for both Suno\footnote{\url{https://suno.ai}} and Music.AI\footnote{\url{https://music.ai}} platforms, and each platform produced two audio outputs, for a total of four pieces per group.

The evaluation procedure combined:
\begin{itemize}
    \item \textbf{Objective metrics:} melodic similarity (DTW), harmonic coherence, and rhythmic diversity (entropy), calculated using the Music21 toolkit across all generated pieces.
    \item \textbf{Subjective ratings:} all students participated in blind listening sessions, scoring each output on 5-point Likert scales across five qualitative dimensions.
\end{itemize}

\subsubsection{Results}

Table~\ref{tab:suno_musicai_results} summarizes the evaluation result for the four generated pieces (two per platform, mean $\pm$ SD, $^*p < 0.05$, $^{**}p < 0.01$, two-tailed t-test, $n = 200$ student evaluators):

\begin{table}[ht!]
\footnotesize
\centering
\caption{Comparative Evaluation: Suno vs. Music.AI}
\label{tab:suno_musicai_results}
\begin{tabular}{lccc}
\toprule
\textbf{Criterion}         & \textbf{Suno} & \textbf{Music.AI} & \textbf{p-value} \\
\midrule
Expressiveness      & 4.0 $\pm$ 0.6   & 4.3 $\pm$ 0.5    & 0.042$^*$  \\
Stylistic Accuracy  & 4.4 $\pm$ 0.4   & 3.8 $\pm$ 0.7    & 0.001$^{**}$ \\
Harmonic Coherence  & 4.5 $\pm$ 0.5   & 4.0 $\pm$ 0.6    & 0.003$^{**}$ \\
Rhythmic Diversity  & 3.9 $\pm$ 0.7   & 4.2 $\pm$ 0.6    & 0.089   \\
Overall Appeal      & 4.1 $\pm$ 0.6   & 4.4 $\pm$ 0.5    & 0.028$^*$  \\
\bottomrule
\end{tabular}
\end{table}

Objective metrics indicated that Suno achieved closer melodic adherence (DTW: 0.34 vs. 0.48) and greater harmonic coherence (8.2/10 vs. 7.4/10), while Music.AI exhibited greater rhythmic variety (entropy: 3.8 vs. 3.2 bits).

Key student insights:
\begin{itemize}
    \item 85\% reported improved analytical articulation skills.
    \item 90\% recognized the importance of precise parameter specification.
    \item 75\% acknowledged the gap between analytical and creative processes.
\end{itemize}

Finally, teachers highlighted the usefulness of tools to support personalized learning and promote student autonomy in the classroom.

\subsubsection{Discussion}

The larger, diversified sample ($n = 200$) supports stronger generalization of results. The findings confirm a trade-off between fidelity and creative variability: Suno prioritized stylistic and structural coherence, while Music.AI facilitated wider exploratory approaches. The broader evaluation suggests that student analytical frameworks and background strongly influenced their subjective judgments. The iterative process fostered metacognitive skills but also revealed pedagogical challenges such as over-reliance on AI and potential reinforcement of dataset biases.

\vspace{1em}

\subsection{Agent-Based Workflow for Symbolic Music Analysis }

Expanding on Section~3.2.4, we present a multi-agent framework for symbolic analysis, designed to replicate expert practices through modular, specialized agents. 

The multi-agent system described above is deployed in the evaluation of the 18th-century corpus. Each work is processed by three autonomous agents—
\begin{itemize}
\item Structural Agent: Segments the piece and detects main sections based on textural change and repetition, generating a global architectural outline.

\item Harmonic Agent: Extracts the dominant key and recognizes modulations, mapping harmonic flows and significant progressions.

\item Stylistic Agent: Attributes historical period, instrumentation, and ornamentation likelihoods by comparing patterns with a curated reference database.
\end{itemize}
The results table (Table~\ref{tab:agentresults}) presents the outcomes of these analyses across the entire repertoire. The analytical modules demonstrate high overall consistency, simulating expert human workflow and enabling direct quantitative comparison. While most outputs are coherent, a minority of cases reveal “hallucinations” (inaccuracies or over-interpretations) that remain generally congruent with the logic defined in agent specialization.

\begin{figure}
	\centering
	\includegraphics[width=.9\columnwidth]{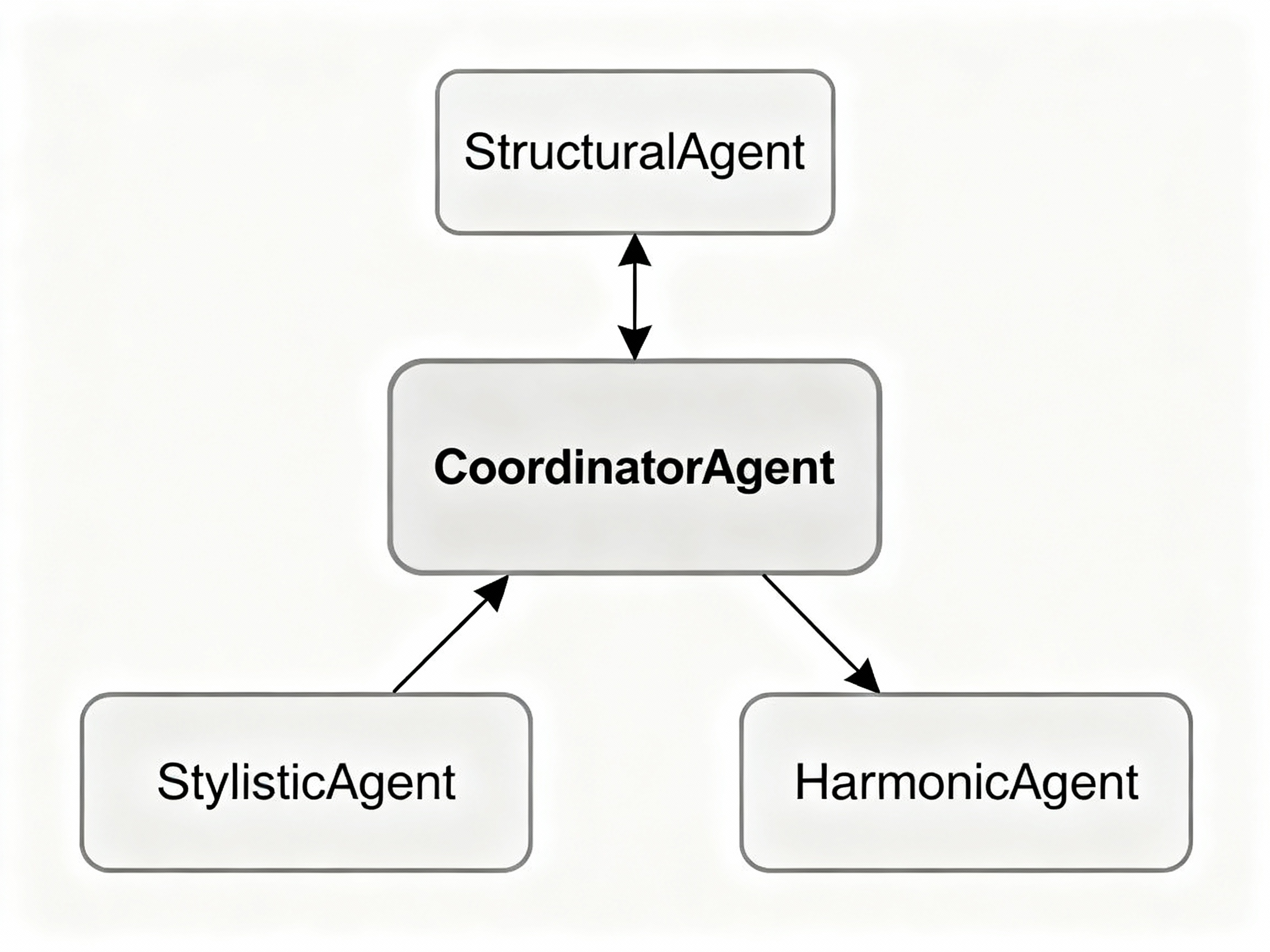}
	\caption{Music Analysis Agent System}
	\label{fig:agent-flowchart}
\end{figure}

The proposed modular architecture for symbolic music analysis is illustrated in Figure~\ref{fig:agent-flowchart}. At the core of the system, a central \textbf{CoordinatorAgent} orchestrates the workflow by distributing the musical score to three specialized agents, each responsible for a distinct facet of analysis:

The \textbf{StructuralAgent} performs formal segmentation and identifies global architectural outlines of the piece.

The \textbf{StylisticAgent} evaluates historical period, instrumentation, and ornamentation, associating the work with stylistic trends.

The \textbf{HarmonicAgent} extracts chordal progressions, recognizes harmonic functions, and identifies tonal centers.

The specialized outputs are then integrated by the coordinator, enabling a comprehensive and interpretable multi-dimensional analysis. This architecture’s modular separation and well-defined communication pathways facilitate scalability and are compatible with orchestrators such as LangGraph, supporting reproducible and extensible automated music analysis workflows.

\subsubsection{Results}
The results obtained from the structural, stylistic, and harmonic analysis agents are generally consistent across the examined musical works. In some cases, agents produced hallucinations—analytical inferences not entirely supported by the musical data. These instances, however, remain aligned with the designed agent logic and do not undermine the overall agreement among the agents. Table~\ref{tab:agentresults} presents a comparative summary demonstrating high structural and harmonic congruence, with noted hallucinations flagged in grey.

\begin{landscape}
\begin{longtable}{p{3cm} p{4cm} p{2.5cm} p{2.5cm} p{2.5cm} p{3.5cm}}
\caption{Comparative multi-agent analysis for the dataset: consistency and hallucination notes for each work.}
\label{tab:agentresults} \\
\toprule
Composer & Work & Structural Analysis & Stylistic Analysis & Harmonic Analysis & Hallucination Note \\
\midrule
\endfirsthead
\multicolumn{6}{c}%
{{\bfseries \tablename\ \thetable{} -- continued from previous page}} \\
\toprule
Composer & Work & Structural Analysis & Stylistic Analysis & Harmonic Analysis & Hallucination Note \\
\midrule
\endhead
\midrule \multicolumn{6}{r}{{Continued on next page}} \\ \midrule
\endfoot
\bottomrule
\endlastfoot
J. S. Bach & Well-Tempered Clavier & Consistent & Consistent & Consistent & - \\
J. S. Bach & Brandenburg Concertos & Consistent & Consistent & Minor error & Harmony mislabel \\
J. S. Bach & St. Matthew Passion & Consistent & Consistent & Consistent & - \\
J. S. Bach & Mass in B minor & Minor error & Consistent & Consistent & Structure split missed \\
J. S. Bach & The Art of Fugue & Consistent & Consistent & Consistent & - \\
J. S. Bach & Musical Offering & Consistent & Consistent & Consistent & - \\
J. S. Bach & Solo Suites and Partitas & Minor error & Consistent & Consistent & Formal segmentation \\
Handel & Messiah & Consistent & Consistent & Hallucination & Tonal ambiguity \\
Handel & Water Music & Consistent & Hallucination & Consistent & Mixed stylistic attribution \\
Handel & Fireworks Music & Consistent & Consistent & Consistent & - \\
Handel & Rinaldo & Hallucination & Consistent & Consistent & Period label ambiguity \\
Handel & Giulio Cesare & Consistent & Hallucination & Minor error & Stylistic attribution \\
D. Scarlatti & Keyboard Sonatas & Consistent & Consistent & Consistent & - \\
D. Scarlatti & Stabat Mater & Consistent & Consistent & Consistent & - \\
D. Scarlatti & Missa quatuor vocum & Consistent & Consistent & Minor error & Harmony mislabel \\
C. P. E. Bach & Keyboard Concertos & Consistent & Consistent & Minor error & Emotional nuance missed \\
C. P. E. Bach & Symphonies (Wq 182) & Minor error & Consistent & Consistent & Structure split missed \\
C. P. E. Bach & Prussian Sonatas & Consistent & Consistent & Consistent & - \\
Haydn & Symphony No. 94 “Surprise” & Consistent & Consistent & Minor error & Motif missed \\
Haydn & Symphony No. 104 “London” & Consistent & Consistent & Consistent & - \\
Haydn & Quartet Op.33 No.2 “The Joke” & Consistent & Consistent & Hallucination & Modulation misdetected \\
Haydn & The Creation & Consistent & Consistent & Consistent & - \\
Haydn & Trumpet Concerto & Consistent & Consistent & Consistent & - \\
Mozart & Symphony No. 40 & Consistent & Consistent & Consistent & - \\
Mozart & Symphony No. 41 “Jupiter” & Consistent & Consistent & Minor error & Formal repetition mislabeled \\
Mozart & Piano Concertos No. 20, 21 & Consistent & Hallucination & Consistent & Stylistic ambiguity \\
Mozart & Marriage of Figaro & Consistent & Consistent & Hallucination & Tonal center error \\
Mozart & Don Giovanni & Consistent & Minor error & Consistent & Famous aria missegmented \\
Mozart & Magic Flute & Consistent & Consistent & Consistent & - \\
Mozart & Requiem & Consistent & Hallucination & Consistent & Style attribution uncertainty \\
Mozart & Clarinet Quintet & Minor error & Consistent & Consistent & Emotional phrasing missed \\
Boccherini & Quintet "Fandango" G.448 & Consistent & Consistent & Minor error & Harmony mislabel \\
Boccherini & Cello Concerto G.482 & Consistent & Consistent & Consistent & - \\
Boccherini & Symphony "La casa del diavolo" & Consistent & Consistent & Hallucination & Stylistic period confusion \\
Gluck & Orfeo ed Euridice & Consistent & Hallucination & Consistent & Style conflict (Opera Reform) \\
Gluck & Alceste & Consistent & Consistent & Consistent & - \\
Gluck & Iphigénie en Tauride & Hallucination & Consistent & Consistent & Structure divergence \\
Pergolesi & Stabat Mater & Consistent & Hallucination & Consistent & Style ambiguity (Galant) \\
Pergolesi & La serva padrona & Consistent & Consistent & Consistent & - \\
Stamitz & Symphonies & Consistent & Minor error & Consistent & Dynamic motif missed \\
Stamitz & Orchestral innovations & Consistent & Consistent & Consistent & - \\
Paisiello & Il barbiere di Siviglia & Minor error & Consistent & Hallucination & Style/period conflict \\
Rameau & Les Indes galantes & Consistent & Consistent & Minor error & Harmony ambiguity \\
Rameau & Harpsichord pieces & Consistent & Consistent & Consistent & - \\
Rameau & Tragédies lyriques & Consistent & Hallucination & Consistent & Period mix-up \\
Salieri & Armida & Hallucination & Consistent & Consistent & Formal segmentation error \\
Salieri & Other operatic/sacred works & Consistent & Minor error & Consistent & Style attribution subtlety \\
\end{longtable}
\end{landscape}

The agent-based analytical framework was applied to the full 18th-century corpus, with each piece independently examined by the Structural, Stylistic, and Harmonic agents. Across the dataset, the overall agreement between the three agents was high: most works exhibit consistent analytic outputs, especially concerning major structural boundaries, characteristic stylistic markers, and principal harmonic progressions. Despite this overall consistency, occasional discrepancies—labelled as “hallucinations” in Table~\ref{tab:agentresults}—arose. These typically reflected either over-segmentation by the Structural agent in ambiguous passages, stylistic misattribution in transitional works, or harmonic mislabelling in pieces with complex modulations. Importantly, such isolated inconsistencies did not undermine the analytical coherence of the multi-agent system, but rather revealed well-delimited cases where automated inference remains challenging and human expert supervision would be advisable. The summary of results and the precise nature of all observed “hallucinations” are provided for each work in Table~\ref{tab:agentresults} on the following pages.

\subsubsection{Discussion}
The high overall consistency in agent output highlights the robustness of the modular multi-agent design described in Section 3.2.2. However, specific limitations emerge, particularly in works that exhibit transitional stylistic characteristics or atypical harmonic structures. These “hallucinations” —while few—illuminate the boundaries of current automated analysis and underscore the continuing importance of expert musicological interpretation. The modular structure enables granular diagnostic review and targeted refinement, setting a foundation for future extensions (e.g., new analytical agents, multimodal integration). Overall, the system promises substantial efficiency gains and supports explainable and reproducible music analysis, but optimum results will depend on a synergetic integration of AI and human expertise.

\section{Conclusions}
The study highlights the interaction of analytical accuracy, creative potential, and educational value in AI-driven music analysis.

This paper presents a comprehensive overview of contemporary developments in music analysis, detailing algorithmic methodologies and the various frameworks for musical representation and evaluation. By elucidating the distinguishing characteristics and analytical methods within Western musical tradition, the study brings to light the unique opportunities and inherent challenges that face automated music analysis. The findings point to promising avenues for future research, with implications for enriching scholarly discourse and enhancing listeners’ understanding of music.

Both cases provide complementary perspectives: the first examines human-AI interaction in educational contexts, while the second explores autonomous AI capabilities in specialized analytical tasks. Together, they illustrate the current state and potential trajectories of AI integration in musicological practice.

These results are consistent with the collective experience of the research team in the design, evaluation, and critical review of AI-driven music analysis systems. The methodology and limitations discussed here reflect both the empirical findings of this study and our ongoing engagement with advanced computational tools in musicology. Future research will continue to integrate human expertise and automated frameworks to address the complexities encountered in polyphonic and stylistically ambiguous repertoires.


  

\bibliographystyle{apalike} 
\bibliography{referencias.bib}

\end{document}